\documentclass[conference,a4paper]{IEEEtran}
\IEEEoverridecommandlockouts

\usepackage{cite}
\usepackage{amsmath,amssymb,amsfonts}
\usepackage{graphicx}
\usepackage{textcomp}
\usepackage{xcolor}
\usepackage{booktabs}
\usepackage{microtype}
\usepackage{pgfplots}
\pgfplotsset{compat=1.18} % Prevents compilation warnings
\usepackage{tikz}

\def\BibTeX{{\rm B\kern-.05em{\sc i\kern-.025em b}\kern-.08em
    T\kern-.1667em\lower.7ex\hbox{E}\kern-.125emX}}

\begin{document}

\title{ECHO: A Locally-Deployable Agentic Health Assistant with Temporal Memory, Safety Guardrails, and Speech Assessment}
\makeatletter
\def\@IEEEauthorblockNfont{\fontsize{11pt}{13pt}\selectfont}
\def\@IEEEauthorblockAfont{\fontsize{10pt}{12pt}\selectfont}
\makeatother

\author{
    \IEEEauthorblockN{
        Abdulkadir K\"{u}l\c{c}e\IEEEauthorrefmark{1}, 
        Alihan Esen\IEEEauthorrefmark{1}, 
        \c{C}a\u{g}la Fikir\IEEEauthorrefmark{2}, 
        Berke Kurt\IEEEauthorrefmark{1},
        Kuzey Arar\IEEEauthorrefmark{2},
        G\"{o}khan Ercan\IEEEauthorrefmark{1},
        Faik Boray Tek\IEEEauthorrefmark{1}
    }
    \IEEEauthorblockA{
        \IEEEauthorrefmark{1}\textit{Dept.\ of Artificial Intelligence and Data Engineering, Istanbul Technical University, Istanbul, Turkey}
    }
    \IEEEauthorblockA{
        \IEEEauthorrefmark{2}\textit{Dept.\ of Computer Engineering, Istanbul Technical University, Istanbul, Turkey}
    }
    \IEEEauthorblockA{
         \{kulce21, esenal22, fikir21, kurtbe21, arar25, ercang, tekfb\}@itu.edu.tr, 
    }
}
% \author{
%   \IEEEauthorblockN{Abdulkadir K\"{u}l\c{c}e, Alihan Esen, \c{C}a\u{g}la Fikir, Berke Kurt}
%   \IEEEauthorblockA{\textit{Dept.\ of Artificial Intelligence and Data Engineering,
%   Istanbul Technical University, Istanbul, Turkey}\\
%   %Advisors: Assoc.\ Prof.\ Dr.\ Faik Boray Tek, Dr.\ G\"{o}khan Ercan}
% }

\maketitle
\thispagestyle{empty}

\begin{abstract}
This paper presents ECHO (Enhanced Care \& Health Observer), a locally-deployable conversational health assistant for long-term chronic care management. ECHO integrates three complementary software modules developed under shared supervision as a unified system. The core module is an agentic chatbot built on a ReAct loop orchestrated via LangGraph, equipped with 17 clinical tools and a temporal knowledge graph for persistent cross-session memory; it achieves a 94.9\% tool-execution pass rate across a 59-scenario benchmark with GPT-5 Mini. A two-stage hybrid safety layer intercepts all incoming queries: a rule-based layer handles explicit crisis signals and jailbreak attempts in under 1ms, while a signed graph neural network (GNN) with APPNP-style propagation classifies boundary cases by clinical intent, achieving 88.8\% accuracy and 90.6\% unsafe recall on a 2,537-query annotated Turkish health dataset while outperforming zero-shot LLM baselines including Llama 3.3 70B. A multimodal speech assessment module combining Whisper acoustic encoding and BERT text encoding with cross-attention fusion estimates emotion, depression, and pain, reaching a mean macro F1 of 0.652. The full system is implemented as a web application that can run entirely on consumer hardware, with no patient data transmitted to external services, supporting compliance with GDPR and KVKK.
\end{abstract}


%──────────────────────────────────────────────────────────────────────────────
\section{Introduction}

Modern chronic care management is episodic by design: patients are expected to
independently execute complex clinical routines between hospital visits with little
continuous support.
All-cause 30-day unplanned readmission rates for chronic disease cohorts reach
21.3\%\,\cite{kaya2018}, directly reflecting a systemic failure in outpatient care
continuity.

Existing digital health tools fail to close this gap for three reasons.
First, LLMs are stateless across sessions: they forget medical history, allergies, and
changing symptoms between conversations.
Second, reminder applications are passive and unable to interpret clinical context or
answer complex follow-up questions.
Third, general-purpose LLMs lack specialized safety mechanisms to intercept dangerous
queries before generating a response.

ECHO (Enhanced Care \& Health Observer) addresses all three limitations through
a unified architecture comprising three integrated modules:
\begin{itemize}
  \item An agentic orchestration layer with a temporal knowledge graph
    maintaining a persistent patient profile across sessions with 17 assistive tools.
  \item A two-stage hybrid guardrail combining deterministic rule-based interception
    for explicit crises with a signed GNN intent classifier for boundary-case queries.
  \item A speech-input assessment module that runs only for voice interactions and injects emotion, depression, and pain estimates into the agent context.
\end{itemize}

%──────────────────────────────────────────────────────────────────────────────
\section{Related Work}
\label{sec:related}

Medical LLMs such as Med-PaLM\,\cite{medpalm} and Med-Gemini\,\cite{corrado2024medgemini}
were fundamentally session-bound: they could answer clinical questions from static
context but had no mechanism for maintaining an enhanced patient history across
separate conversations.
RAG\,\cite{lewis2020rag} addressed this partially by retrieving patient-specific
records at query time, but the append-only structure of vector databases introduces
temporal hallucination when clinical facts change. Latimer et al.\ \cite{latimer2025hindsight} claim state-of-the-art performance for \textit{Hindsight}, which organizes knowledge into typed fact hierarchies and fuses multi-strategy retrieval with Reciprocal Rank Fusion.
LongMemEval\,\cite{wu2024longmemeval} benchmarks show that commercial LLMs frequently
fail to overwrite obsolete information, posing a direct safety risk in clinical
contexts where outdated dosage records could inform dangerous advice.
Unlike these session-bound or append-only approaches, ECHO combines a structured
SQLite layer for transactional health data with a Hindsight temporal knowledge graph
for unstructured longitudinal memory, allowing clinical facts to be superseded rather
than accumulated, preserving a safe and consistent patient history across sessions.

\subsection{Safety Classification for Health Queries}
Keyword-based guardrails are easily avoided by rephrasing and generate high
false-positive rates on clinical terms that are hazardous in one context but benign
in another; yet they remain useful for unambiguous signals where neural inference
latency is unacceptable.
Llama Guard\,\cite{inan2023llamaguard} frames safety as instruction-following via a
7B LLM, offering taxonomy flexibility but at inference cost impractical for a
low-latency pre-agent filter.
Abercrombie and Rieser\,\cite{abercrombie2022} show that clinical severity cannot be
reliably inferred from surface language alone, precisely the regime where a learned
classifier is necessary beyond simple rules. Approximate personalized propagation of neural predictions
(APPNP)\, \cite{gasteiger2019} and signed Graph Neural Networks (GNNs)\,\cite{derr2018sgcn} have been applied
to other NLP tasks but not previously to health query safety classification.
% Unlike Llama Guard, which requires a full 7B-parameter forward pass at every
% inference call, ECHO's signed GNN classifier requires only a sentence encoder
% embedding and a small MLP applied to a pre-computed $\sim$8\,MB nearest-neighbor
% index, achieving comparable or superior accuracy at orders-of-magnitude lower
% computational cost.

\subsection{Speech-Based Health \& Mood Assessment}

Recent speech foundation models such as wav2vec\,2.0\,\cite{baevski2020wav2vec}
and Whisper\,\cite{radford2022whisper} provide strong acoustic representations
that can be adapted to downstream clinical tasks, while BERT\,\cite{devlin2019bert}
enables transcript-level semantic modeling. Together, these encoders make it
possible to analyze both how a user speaks and what the user says.

Most existing speech-health systems, however, are designed for a single target
condition. Prior work has studied depression detection\,\cite{liu2024depression},
Parkinson's disease detection\,\cite{purohit2025}, and voice-based pain
assessment\,\cite{borna2023} as separate tasks, typically training one model per
condition. This single-task framing limits their usefulness in conversational
care settings, where a single voice interaction may contain multiple
health-relevant signals. Multi-task learning\,\cite{caruana1997} offers a
natural alternative by allowing related tasks to share a common representation
while preserving task-specific prediction heads.

Unlike these single-task systems, ECHO's speech module estimates emotion, depression, and pain from the same voice interaction, allowing the agentic pipeline to receive a richer passive health signal without additional user input.

%──────────────────────────────────────────────────────────────────────────────
\section{System Architecture}
\label{sec:arch}

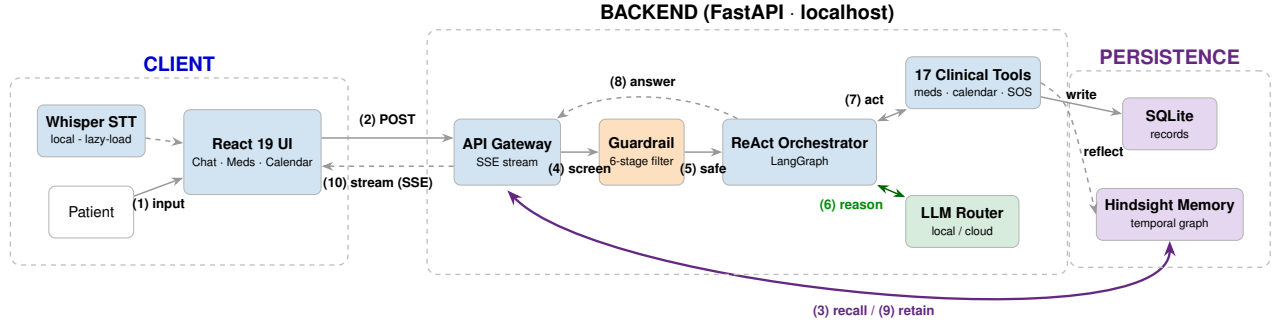
\begin{figure*}[htbp]
  \centering
  % Set explicit width and height, and use keepaspectratio=false to allow vertical stretching
  \resizebox{\textwidth}{!}{%
    \usetikzlibrary{positioning, fit, shapes.geometric, arrows.meta, calc, backgrounds}

\definecolor{clientblue}{RGB}{230, 242, 250}
\definecolor{backendgray}{RGB}{245, 247, 250}
\definecolor{persistpurple}{RGB}{243, 235, 245}

\definecolor{nodeblue}{RGB}{210, 228, 242}
\definecolor{nodegreen}{RGB}{215, 236, 220}
\definecolor{nodeorange}{RGB}{253, 224, 191}
\definecolor{nodepurple}{RGB}{230, 215, 240}
\definecolor{textpurple}{RGB}{100, 40, 130}

\definecolor{textgreen}{RGB}{1, 139, 10}

\begin{tikzpicture}[
    font=\sffamily\small, % Increased global base font size for better legibility
    >=Stealth,
    block/.style={
        rectangle, draw=gray!60, fill=nodeblue, minimum height=1.1cm, minimum width=1.8cm, 
        rounded corners=4pt, align=center, line width=0.6pt, inner sep=5pt
    },
    container/.style={
        rectangle, draw=gray!50, dashed, rounded corners=6pt, line width=0.8pt, inner sep=16pt
    },
    arrow/.style={->, draw=gray!80, line width=0.9pt},
    dashedarrow/.style={->, draw=gray!80, dashed, line width=0.9pt},
    purplearrow/.style={<->, draw=textpurple, line width=1.4pt},
 % === REPLACE THIS LINE IN YOUR CODE ===
    stepnum/.style={font=\sffamily\bfseries\footnotesize} 
    % ======================================
]

    % =========================================================================
    % CLIENT SIDE
    % =========================================================================
    \node[block] (whisper) {\textbf{Whisper STT}\\\scriptsize local - lazy-load};
    \node[block, below=0.6cm of whisper, fill=white] (patient) {Patient};
    \node[block, right=0.8cm of whisper, minimum height=1.8cm, yshift=-0.45cm, minimum width=2.2cm] (reactui) {\textbf{React 19 UI}\\\scriptsize Chat $\cdot$ Meds $\cdot$ Calendar};

    \path[dashedarrow] (whisper) -- (reactui);
    \path[arrow] (patient) -- node[below=0.1cm, stepnum] {(1) input} (reactui);

    \node[container, fit=(whisper) (patient) (reactui), label={[text=blue!80!black, font=\sffamily\bfseries\large]above:CLIENT}] (clientbox) {};

    % =========================================================================
    % BACKEND SIDE
    % =========================================================================
    \node[block, right=2.8cm of reactui, minimum height=1.4cm, minimum width=2.2cm] (apigateway) {\textbf{API Gateway}\\\scriptsize SSE stream};
    \node[block, right=0.8cm of apigateway, fill=nodeorange, minimum height=1.4cm] (guardrail) {\textbf{Guardrail}\\\scriptsize 6-stage filter};
    \node[block, right=0.8cm of guardrail, minimum height=1.4cm, minimum width=2.4cm] (reactorch) {\textbf{ReAct Orchestrator}\\\scriptsize LangGraph};
    
    \node[block, above right=0.2cm and 0.6cm of reactorch, minimum width=2.4cm] (tools) {\textbf{17 Clinical Tools}\\\scriptsize meds $\cdot$ calendar $\cdot$ SOS};
    \node[block, below right=0.2cm and 0.6cm of reactorch, fill=nodegreen, minimum width=2.4cm] (llmrouter) {\textbf{LLM Router}\\\scriptsize local / cloud};

    \path[arrow] (apigateway) -- node[below=0.1cm, stepnum] {(4) screen} (guardrail);
    \path[arrow] (guardrail) -- node[below=0.1cm, stepnum] {(5) safe} (reactorch);
    \path[arrow, <->] (reactorch) -- node[above left=0.05cm, stepnum] {(7) act} (tools);
    \path[arrow, <->, draw=green!40!black] (reactorch) -- node[below left=0.05cm, stepnum, text=textgreen] {(6) reason} (llmrouter);

    \node[container, fit=(apigateway) (guardrail) (reactorch) (tools) (llmrouter), label={[text=black, font=\sffamily\bfseries\large]above:BACKEND (FastAPI $\cdot$ localhost)}] (backendbox) {};

    % =========================================================================
    % PERSISTENCE SIDE
    % =========================================================================
    \node[block, right=5.2cm of reactorch, fill=nodepurple, yshift=0.6cm, minimum width=2.0cm] (sqlite) {\textbf{SQLite}\\\scriptsize records};
    \node[block, below=0.8cm of sqlite, fill=nodepurple, minimum width=2.6cm] (hindsight) {\textbf{Hindsight Memory}\\\scriptsize temporal graph};

    \node[container, fit=(sqlite) (hindsight), label={[text=textpurple, font=\sffamily\bfseries\large]above:PERSISTENCE}] (persistbox) {};

    % Connect Tools to SQLite and Memory with annotations
    \path[arrow] (tools) -- node[above, stepnum] {write} (sqlite);
    \path[dashedarrow] (tools.east) .. controls ($(tools.east)+(0.6,-0.4)$) .. node[below right, stepnum, pos=0.7] {reflect} (hindsight.west);

    % =========================================================================
    % COUPLINGS & TIMESTAMPS
    % =========================================================================
    
    % (2) POST and (10) stream (SSE)
    \path[arrow] ([yshift=0.3cm]reactui.east) -- node[above=0.1cm, stepnum] {(2) POST} ([yshift=0.3cm]apigateway.west);
    \path[dashedarrow] ([yshift=-0.3cm]apigateway.west) -- node[below=0.1cm, stepnum, xshift=-0.2cm] {(10) stream (SSE)} ([yshift=-0.3cm]reactui.east);

    % (8) answer dashed path overhead
    \path[dashedarrow] (reactorch) .. controls ($(reactorch.north west)+(-0.8,0.6)$) and ($(apigateway.north east)+(0.8,0.6)$) .. node[above=0.05cm, stepnum] {(8) answer} (apigateway);

    % =========================================================================
    % BACKGROUND LAYER: (3) recall / (9) retain route underneath the boxes
    % =========================================================================
\begin{pgfonlayer}{background}
        % Sweeps cleanly downward from Hindsight Memory, underneath the components, up back to API Gateway
        \path[purplearrow] (hindsight.south) .. controls ($(hindsight.south)+(0,-2.2)$) and ($(apigateway.south)+(2.0,-2.2)$) .. 
            node[below=0.15cm, stepnum, text=textpurple, pos=0.5] { (3) recall / (9) retain} 
            ([yshift=-0.1cm]apigateway.south);
    \end{pgfonlayer}

\end{tikzpicture}%
  }
  \caption{End-to-end ECHO system architecture. Numbered arrows trace the
    request lifecycle: (1)~user input, (2)~POST to API Gateway,
    (3)~cross-session context recall from Hindsight, (4)~guardrail screening,
    (5)~safe input to LangGraph ReAct orchestrator, (6)~LLM reasoning via
    LiteLLM, (7)~clinical tool execution against SQLite, (8)~response
    generation, (9)~asynchronous memory retention, (10)~SSE streaming back to
    the React frontend. Whisper speech-to-text runs locally on the client.}
  \label{fig:arch}
\end{figure*}

ECHO's React frontend communicates with a FastAPI backend that orchestrates the
full pipeline.
The request lifecycle proceeds as follows: the user message is posted to
the API Gateway, which first queries the Hindsight memory engines for relevant
cross-session context, then forwards the enriched input to the hybrid guardrail.
If the guardrail passes the query, it reaches the LangGraph agentic loop, which
reasons, invokes tools against SQLite, and generates a response streamed back via
Server-Sent Events.
After the stream completes, a background coroutine asynchronously retains the new
turn in Hindsight.
When the user submits voice input, the audio is transcribed locally and simultaneously analyzed by the speech assessment module; the resulting emotion, depression, and pain estimates are injected into the agent's context alongside the transcript.
All components are designed to support fully local operations on the user’s machine, with no personally identifiable
health data transmitted to external services, ensuring compliance with GDPR and KVKK
regulations (Fig.\,\ref{fig:arch}).

\subsection{Agentic Orchestration and Temporal Memory}

The dialogue engine implements a Reason and Act (ReAct)\cite{yao2023react} loop via a LangGraph state
graph.
The conversation history is maintained in a message-state schema, persisted across
requests by a memory-saver checkpointer keyed to the active session thread.
The graph connects two operational nodes: an \emph{LLM Call Node} that resolves the
system prompt (dynamically appending the current timezone-aware timestamp for
temporal grounding, critical for rejecting future-dated medication logs or detecting
overdue doses) and invokes the model through a LiteLLM gateway providing a unified
interface for local Ollama models and cloud APIs; and a \emph{Tool Node} that
receives tool calls from the latest LLM message, executes the referenced Python
functions against the local SQLite database, and appends operational results to the
message state.
The loop terminates when the LLM produces a message without tool calls.
Independent tool calls within a single turn are parallelized, reducing latency for
compound requests such as simultaneously logging a symptom and marking a dose taken.

The 17 assistive tools are organized into four functional categories:
\begin{itemize}
  \item \emph{Medication management}: add, remove, list, log intake with adherence
    tracking, undo log, and duplicate detection before insertion.
  \item \emph{Calendar scheduling}: add and remove appointments; expand recurring
    event rules (daily, weekly, bi-weekly, monthly) into chronological occurrences.
  \item \emph{Symptom logging}: record severity scores on predefined numerical scales,
    categorized by clinical tags (tremor, rigidity, freezing, nausea).
  \item \emph{Emergency protocols}: store caregiver contacts, retrieve and display
    them on crisis detection, and trigger deep longitudinal memory reflection via
    the \texttt{reflect} tool for cross-session health summaries.
\end{itemize}
% After each turn, a background FastAPI coroutine asynchronously ships the conversation
% to the Hindsight knowledge graph, decoupling memory ingestion latency from
% user-perceived response time and allowing continuous interaction during synthesis.

\subsection{Hindsight Temporal Knowledge Graph}

Long-term memory is provided by Hindsight, deployed as a local Docker container.
The knowledge graph organizes information into three typed memory networks:
\begin{itemize}
  \item \emph{World Facts}: objective, static entity-relation assertions (e.g.,
    ``patient is allergic to Penicillin'').
  \item \emph{Experience Facts}: chronological records of agent-patient interactions
    (e.g., ``patient logged tremor severity 8 at 10:15\,AM on March\,11'').
  \item \emph{Observations}: deduplicated, evidence-grounded beliefs automatically
    consolidated from raw facts by a background process.
    Each observation carries a \emph{proof count} recording the number of independent
    facts corroborating it.
\end{itemize}
During each asynchronous consolidation cycle, the background process extracts atomic
tuples from conversation turns, resolves entity aliases, and wires new facts to
existing graph nodes via temporal, semantic, causal, and co-occurrence edges.
When a fact contradicts an existing observation, the prior belief is superseded with
a timestamped revision rather than deleted, preserving the clinical history for audit.

Retrieval is coordinated by the TEMPR engine executing four parallel strategies:
semantic vector search via HNSW-indexed pgvector, BM25 keyword matching, graph-based
link expansion, and temporal range filtering, all fused through Reciprocal Rank Fusion.
% \begin{equation}
%   \mathrm{RRF\_score}(d)
%     = \sum_{r \in \text{rankers}} \frac{1}{k + \mathrm{rank}_r(d)},\quad k=60
% \end{equation}
Top candidates are passed through a neural cross-encoder reranker before injection
into the LLM context.
A novel \emph{Proof-Count Boost} then up-ranks observations corroborated by many
independent facts, using a logarithmic normalization applied as a multiplicative
score adjustment.
% \begin{equation}
%   S_{\mathrm{final}} = S_{\mathrm{base}}
%     \times \bigl(1 + \alpha\,(\mathrm{clamp}(0.5{+}\tfrac{\ln C}{10},0,1) - 0.5)\bigr),
%     \;\alpha{=}0.1
% \end{equation}

This caps the maximum score lift at 5\% for highly corroborated observations
($C\!\geq\!149$), preventing frequent but low-relevance memories from polluting the
context window regardless of frequency.
% The \texttt{retain\_mission} parameter restricts fact extraction to clinically
% relevant content (medications, symptoms, appointments), filtering out
% conversational noise.
Entity creation is restricted to predefined clinical taxonomy templates.
Temporal decay applies Ebbinghaus-inspired linear forgetting \cite{ebbinghaus1913memory}, ensuring older
unrepeated facts naturally lose retrieval priority unless supported by new evidence.

\subsection{Hybrid Safety Guardrail Pipeline}

Every message is intercepted by a two-stage pipeline before reaching the LLM.

\paragraph{Stage\,1 --- Rule-Based Layer}
Three sequential regex checks run in under 1\,ms:
(1)~\emph{Jailbreak/injection}: structural patterns for role-override and prompt
extraction attempts;
(2)~\emph{Explicit crisis}: bi-contextual matching requiring co-occurrence of a
medication token with a self-harm keyword within a bounded window;
(3)~\emph{Off-topic filter}: requests structurally unrelated to health (code
generation, finance, etc.).
Matches are blocked immediately; the LLM is never invoked.
Crisis matches surface the patient's emergency contacts; jailbreak and off-topic
matches receive a brief refusal.

\paragraph{Stage\,2 --- Signed GNN Classifier}
Queries passing the rule-based layer are forwarded to the signed GNN, which
handles the harder class of \emph{boundary cases}: clinically dangerous queries
that contain no explicit harmful language.
It classifies each query into a nine-class intent taxonomy (\textit{benign
general, medication related, mild emotional stress, chronic condition, non-acute
injury, acute medical risk, harmful treatment, manipulation, self-harm
risk}) and simultaneously produces a binary safe/unsafe label.

Queries are encoded by sentence-transformer model \texttt{paraphrase-multilingual-mpnet-base-v2}
(768-dim) and passed through a shared 512-dim encoder feeding a safety head and
an intent head.
A $k$-NN graph ($k\!=\!15$, cosine distance) over 2,029 training embeddings
produces positive edges (same-class, similarity $\geq\!0.75$; 3,742 edges) and
negative edges at six empirically identified confusion boundaries (cross-class,
similarity $\geq\!0.70$; 1,625 directed edges from unsafe to safe nodes).

Positive propagation follows APPNP ($\alpha\!=\!0.2$, $K\!=\!2$):
\begin{equation}
  H^{(t+1)} = (1{-}\alpha)\,A_{\mathrm{pos}}\,H^{(t)} + \alpha\,H_0
\end{equation}
A negative correction delta repels boundary-case embeddings from confusable
cross-class representations:
\begin{equation}
  H_{\mathrm{final}} = H_{\mathrm{pos}}
    - \lambda_{\mathrm{neg}}\,\alpha_{\mathrm{neg}}
      (A_{\mathrm{neg}}\,H_{\mathrm{pos}} - H_{\mathrm{pos}})
\end{equation}
Training uses a five-term loss: safety cross-entropy with hard example mining
(weight 15.0 for missed unsafe), intent cross-entropy with risk-severity weights
(\textit{self-harm risk}: 10.0), pull and push contrastive losses, and a
propagation consistency term.
% At deployment, only the trained weights and a pre-computed ${\sim}$8\,MB
% nearest-neighbor embedding index are required; the raw training dataset is not
% needed at inference time.

\paragraph{Intent-Aware Response Routing}
Unsafe GNN predictions trigger intent-specific responses: \textit{self-harm risk}
surfaces emergency contacts with a supportive message; \textit{acute medical risk}
advises immediate emergency care; \textit{harmful treatment} recommends physician
consultation; \textit{manipulation} receives a silent generic refusal to prevent
adversarial feedback.
Safe predictions are forwarded to LangGraph with the intent label in context.

\subsection{Speech Assessment in the Voice Input Pipeline}

The speech assessment module provides a passive health signal during voice-based interactions.
When the user speaks through the browser’s push-to-talk interface, the audio is transcribed locally with Whisper and analyzed before the message enters the chat pipeline.

The module estimates three complementary health-relevant dimensions: emotional state
(5-class: neutral, happy, sad, angry, fear), depression status (binary), and pain
status (binary).
The estimated scores are injected into the
agent's system context alongside the transcribed text, allowing the LangGraph
orchestrator to proactively adjust its response tone and suggest the
\texttt{log\_symptom} tool when elevated pain or distress is detected.

The model combines acoustic and textual information using two pretrained encoders. Whisper-base is used as the audio encoder to capture prosodic and acoustic cues such as tone, rhythm, and vocal tension, while BERT encodes the Whisper-generated transcript to capture semantic information. A cross-attention fusion block then aligns what the user says with how it is spoken. The resulting fused representation is passed to three independent classification heads for emotion, depression, and pain prediction.

Training combines five corpora under partial-label learning: IEMOCAP\,\cite{busso2008iemocap}, CREMA-D\,\cite{cao2014cremad}, and
RAVDESS\,\cite{livingstone2018ravdess} for emotion; DAIC-WOZ\,\cite{gratch2014daicwoz} for depression; and TAME Pain\,\cite{dao2025tamepain}
for pain.
Because each dataset provides labels for only one task, each sample updates only the relevant prediction head while still contributing to the shared multimodal representation.
To prevent the larger emotion datasets from dominating training, task-balanced sampling is used. During fine-tuning, the upper Whisper layers and fusion/classification heads are updated, while BERT remains frozen to reduce overfitting on the smaller clinical datasets.

\section{Experimental Results}
\label{sec:results}

% \subsection{Agentic Tool Execution}

The benchmark covers 59 scenarios and 110 turns across nine clinical categories.
Table\,\ref{tab:tools} reports Pass Rate and argument-level F1 for selected models:
% Fig.\,\ref{fig:passrate} provides a visual comparison across all evaluated models.

% \begin{figure}[htbp]
%   \centering
%   \input{result_bar}
%   \caption{Tool-execution pass rates across 13 LLMs on the 59-scenario
%     benchmark. The dashed line marks the $\geq$90\% design target.
%     GPT-5 Mini (primary deployment model, blue) and Gemma\,4\,26B
%     (self-hostable, green) are highlighted.}
%   \label{fig:passrate}
% \end{figure}

\begin{table}[htbp]
  \caption{Tool Execution Benchmark (59-Scenario Suite)}
  \label{tab:tools}
  \centering\small
  \begin{tabular}{lcccc}
    \toprule
    \textbf{Model} & \textbf{Pass} & \textbf{F1} &
      \textbf{TR Pass} & \textbf{TR F1} \\
    \midrule
    GPT-OSS 120B       & 96.61\% & 0.980 & ---     & ---   \\
    GPT-5 Mini         & 94.92\% & 0.977 & 92.86\% & 0.976 \\
    Gemma 4 26B IT     & 89.83\% & 0.951 & ---     & ---   \\
    GPT-5 Nano         & 88.14\% & 0.915 & 92.86\% & 0.929 \\
    Qwen 3 32B         & 86.44\% & 0.897 & ---     & ---   \\
    GPT-5.4 Nano       & 76.27\% & 0.818 & 50.00\% & 0.673 \\
    Gemini 2.5 Fl.Lite & 47.46\% & 0.548 & ---     & ---   \\
    \bottomrule
  \end{tabular}
\end{table}

GPT-5 Mini achieves the best cost-accuracy trade-off (94.92\%, F1 0.977), meeting
the $\geq\!90\%$ design target.
High-capacity commercial models consistently exceed 94\% pass rate, while
medium-sized open models remain above 86\%, making them viable for self-hosted
edge deployments.
% Turkish cross-lingual sensitivity is significant: GPT-5 Mini maintains
% near-identical performance (92.86\%), while GPT-5.4 Nano drops to 50\% due to
% agglutinative morphology causing malformed parameter extraction on medication names
% suffixed with Turkish case markers.
End-to-end latency per turn ranges from 4.82\,s (GPT-5.4 Nano) to 8.78\,s
(GPT-5 Mini); SSE streaming keeps Time to First Token under 3\,s across all models.
Failure mode analysis of GPT-5 Mini's three failures reveals: (1) the model sought
clarification instead of invoking the tool when appointment time was unspecified;
(2) a complementary \texttt{mark\_medication\_taken} call was omitted in a week-long
timeline scenario; (3) a Turkish list-medications query triggered redundant invocation
of both \texttt{list\_medications} and \texttt{get\_todays\_medications}.

Cross-session memory was validated through a one-week patient simulation spanning
10 conversation threads.
Unstructured facts absent from the SQLite schema (patient name, neurologist identity,
and lifestyle preferences) were consistently recalled without re-prompting.
The \texttt{reflect} tool synthesized cross-session patterns such as ``the tremor
on March\,11 came after a late morning dose; the fall on March\,13 was on a day with
no doses logged'', a pattern invisible within any single conversation.
The memory layer also surfaced a neurologist appointment that the tool's
7-day window had omitted.

\subsection{Safety Classifier}

Table\,\ref{tab:ablation} shows the ablation on 508 held-out Turkish queries.
The rule-based layer provides deterministic coverage at zero inference cost;
the signed GNN handles the boundary cases that pass it.
Positive propagation yields the largest single gain (+4.8\,pp unsafe recall);
the full model recovers this gain while adding intent-aware representations,
achieving 88.8\% accuracy and 90.6\% unsafe recall.

\begin{table}[htbp]
  \caption{Safety Classifier Ablation ($n=508$)}
  \label{tab:ablation}
  \centering\small
  \begin{tabular}{lccc}
    \toprule
    \textbf{Configuration} & \textbf{Acc.} & \textbf{Safe R.} & \textbf{Unsafe R.} \\
    \midrule
    Baseline MLP                     & 0.874 & 0.890 & 0.858 \\
    +\,Pos.\ Prop.\ ($\lambda$=0.8)  & 0.886 & 0.866 & 0.906 \\
    +\,Neg.\ Delta ($\lambda_n$=0.5) & 0.874 & 0.878 & 0.870 \\
    +\,Intent Loss                   & 0.876 & 0.862 & 0.890 \\
    Full Model (+\,Pull+Push)        & \textbf{0.888} & 0.870 & \textbf{0.906} \\
    \bottomrule
  \end{tabular}
\end{table}

Table\,\ref{tab:llm} compares the full model against zero-shot baselines.
The proposed system outperforms all baselines in accuracy and balanced accuracy
at orders-of-magnitude lower inference cost.
Llama\,3.3\,70B approaches the system (0.855 balanced accuracy) without
task-specific training, yet requires a 70B-parameter forward pass per query;
ECHO requires only a sentence encoder and a small MLP over a pre-computed 8\,MB index.
False negatives share a common pattern: clinically severe queries phrased as
routine questions, where graph propagation recovers boundary cases with
informative neighbors but cannot recover isolated rare symptom presentations.

\begin{table}[htbp]
  \caption{LLM Baseline Comparison (Zero-Shot)}
  \label{tab:llm}
  \centering\small
  \begin{tabular}{lccc}
    \toprule
    \textbf{Model} & \textbf{Acc.} & \textbf{Safe R.} & \textbf{Unsafe R.} \\
    \midrule
    Llama 3.1 8B   & 0.695 & 0.445 & 0.943 \\
    Qwen3 32B      & 0.813 & 0.843 & 0.785 \\
    Gemini 2.5 FL  & 0.848 & 0.785 & 0.908 \\
    Llama 3.3 70B  & 0.856 & 0.854 & 0.856 \\
    Proposed       & \textbf{0.888} & \textbf{0.870} & \textbf{0.906} \\
    \bottomrule
  \end{tabular}
\end{table}

\subsection{Multimodal Speech Assessment}

Table\,\ref{tab:speech} reports macro F1 scores on participant-level held-out splits for the speech assessment module. The baseline is an audio-only Whisper model trained without the BERT text branch and without cross-attention fusion. It uses the same general training setup and evaluation protocol as the final model, making it a controlled reference for measuring the contribution of multimodal fusion. The final model adds frozen BERT transcript embeddings and a cross-attention fusion block that combines linguistic information with Whisper acoustic representations.

\begin{table}[htbp]
  \caption{Speech Assessment Results (Macro F1)}
  \label{tab:speech}
  \centering\small
  \begin{tabular}{lccc}
    \toprule
    \textbf{Task} & \textbf{Baseline} & \textbf{Final} & $\Delta$ \\
    \midrule
    Emotion (5-class)   & 0.698 & 0.732 & +0.034 \\
    Pain (binary)       & 0.596 & 0.685 & +0.089 \\
    Depression (binary) & 0.528 & 0.538 & +0.010 \\
    \midrule
    Mean                & 0.607 & \textbf{0.652} & +0.045 \\
    \bottomrule
  \end{tabular}
\end{table}

The final model improves mean macro F1 from 0.607 to 0.652, indicating that adding BERT-based transcript features and cross-attention fusion provides a clear benefit over the Whisper-only baseline. The largest gain is observed in pain detection, suggesting that pain-related speech benefits from combining acoustic cues with transcript-level information. Emotion recognition also improves moderately.

Depression detection remains the most difficult task, with only a small improvement over the baseline. This suggests that short utterance-level modeling is limited for depression screening, which likely requires longer conversational context. Overall, the results support the multimodal fusion design while identifying depression detection as the main direction for future improvement.

%──────────────────────────────────────────────────────────────────────────────
\section{Discussion and Conclusion}
\label{sec:conclusion}

% The three modules address distinct but complementary failure modes of existing
% health AI systems.
% The Hindsight temporal memory resolves statelessness by persisting facts that no
% structured schema can capture across sessions.
% The hybrid guardrail resolves the safety gap in a computationally stratified way:
% deterministic rules handle clear cases cheaply, while the signed GNN handles
% boundary cases at $\sim$20\,ms per query.
% The speech module provides a passive signal channel complementing the explicit
% text-based interaction without requiring the patient to type. In particular, it extends voice input beyond transcription by estimating emotion, pain, and depression-related signals before the message reaches the agent. These estimates provide additional context for agent behavior, such as adjusting response tone or suggesting symptom logging when distress or pain is detected.
% Taken together, memory ensures the agent has accurate patient context; the
% guardrail ensures that context is never used to generate harmful responses;
% and the speech module provides continuous health monitoring through voice.

This paper presented ECHO, a unified chronic care software system integrating
three coupled modules into a locally-deployable pipeline.
The key engineering contribution is the architectural composition: a stateful
agentic core with persistent cross-session memory, a computationally stratified
safety layer, and a passive speech-assessment channel, each deployable on
consumer hardware with no patient data transmitted externally.

Four concrete directions remain.
First, the rule-based crisis lexicon requires adversarial paraphrase
augmentation.
Second, signed GNN recall on rare symptom presentations requires targeted
dataset expansion.
Third, depression detection requires more diverse and task-specific training data, since short speech segments provide limited evidence for reliable screening.
Fourth, ECHO's modular architecture is designed to support extensibility through
the Model Context Protocol (MCP): each clinical tool can be exposed as an
MCP-compliant server, allowing third-party applications (electronic health
record systems, wearable sensor platforms, or pharmacy databases) to integrate
with ECHO through a standardized interface.
This would enable end users and clinicians to configure the assistant's
capabilities to their specific care workflows without modifying the core system.
Longer-term goals include wearable sensor integration, native mobile deployment,
and clinical validation.

% \section*{Acknowledgment}
%  Computational resources
% were provided by the Turkish National High-Performance
% Computing Center (UHeM, Project Number ??)

%──────────────────────────────────────────────────────────────────────────────
\bibliographystyle{IEEEtran}
\bibliography{references}

\end{document}